\documentclass[10pt,twocolumn,letterpaper]{article}

\usepackage[pagenumbers]{cvpr} % To force page numbers, e.g. for an arXiv version

\usepackage[dvipsnames]{xcolor}

\definecolor{cvprblue}{rgb}{0.21,0.49,0.74}
\usepackage[pagebackref,breaklinks,colorlinks,citecolor=cvprblue]{hyperref}
\usepackage{graphicx}
\usepackage{amsmath}
\usepackage{amssymb}
\usepackage{booktabs}
\usepackage{multirow}
\usepackage{xcolor}
\usepackage{tabularx}
\usepackage{colortbl}  % colorful tables
\usepackage{array}
\usepackage{siunitx} %typeset numbers for \num
\usepackage{tcolorbox}  % for prompt design
\usepackage{listings}
\usepackage{setspace}
\usepackage{marvosym}   % for letter

\definecolor{codegreen}{rgb}{0,0.6,0}
\definecolor{codegray}{rgb}{0.5,0.5,0.5}
\definecolor{codepink}{RGB}{252, 142, 172}
\definecolor{codepurple}{rgb}{0.58,0,0.82}
\definecolor{backcolour}{RGB}{245,245,245}
\lstdefinestyle{mystyle}{
    backgroundcolor=\color{backcolour},   
    commentstyle=\color{magenta},
    keywordstyle=\color{blue},
    numberstyle=\tiny\color{codegray},
    stringstyle=\color{codepurple},
    basicstyle=\fontfamily{\ttdefault}\footnotesize,
    breakatwhitespace=false,         
    breaklines=true,                 
    keepspaces=true,    
    frame=single,
    numbersep=5pt,                  
    showspaces=false,                
    showstringspaces=false,
    showtabs=false,                  
    tabsize=2,
    classoffset=1, % starting new class
    keywordstyle=\color{violet},
    classoffset=0,
}
\lstdefinelanguage{JavaScript}{
  keywords={typeof, new, true, false, catch, function, return, null, catch, switch, var, if, in, while, do, else, case, break},
  keywordstyle=\color{blue}\bfseries,
  ndkeywords={class, export, boolean, throw, implements, import, this},
  ndkeywordstyle=\color{darkgray}\bfseries,
  identifierstyle=\color{black},
  sensitive=false,
  comment=[l]{//},
  morecomment=[s]{/*}{*/},
  commentstyle=\color{purple}\ttfamily,
  stringstyle=\color{red}\ttfamily,
  morestring=[b]',
  morestring=[b]"
}

\def\paperID{9300} % *** Enter the Paper ID here
\def\confName{CVPR}
\def\confYear{2024}

\title{ThinkBot: Embodied Instruction Following with Thought Chain Reasoning}

\author{
    Guanxing Lu$^{1}$, 
    Ziwei Wang$^{2}$\textsuperscript{\Letter}\thanks{\Letter~ Corresponding author.},
    Changliu Liu$^{2}$,
    Jiwen Lu$^{3}$,
    Yansong Tang$^{1}$\\
    $^1$Tsinghua Shenzhen International Graduate School, Tsinghua University\\
    $^2$Carnegie Mellon University~~~
    $^3$Department of Automation, Tsinghua University\\
    {\tt\small \{lgx23@mails.,lujiwen@,tang.yansong@sz.\}tsinghua.edu.cn}~~
    {\tt\small \{ziweiwa2@,cliu6@\}andrew.cmu.edu}\\
    {\small \url{https://guanxinglu.github.io/thinkbot/}
    }
}

\usepackage{mwe} % qualitative examples
\usepackage{calc} % qualitative examples
\usepackage{pifont} % checkmark in ablation

\definecolor{Gray}{gray}{0.85}
\newcommand{\method}{\mbox{{ThinkBot}}\xspace}

\begin{document}
\maketitle
\begin{abstract}

    Embodied Instruction Following (EIF) requires agents to complete human instruction by interacting objects in complicated surrounding environments. Conventional methods directly consider the sparse human instruction to generate action plans for agents, which usually fail to achieve human goals because of the instruction incoherence in action descriptions. On the contrary, we propose ThinkBot that reasons the thought chain in human instruction to recover the missing action descriptions, so that the agent can successfully complete human goals by following the coherent instruction. Specifically, we first design an instruction completer based on large language models to recover the missing actions with interacted objects between consecutive human instruction, where the perceived surrounding environments and the completed sub-goals are considered for instruction completion. Based on the partially observed scene semantic maps, we present an object localizer to infer the position of interacted objects for agents to achieve complex human goals. Extensive experiments in the simulated environment show that our ThinkBot outperforms the state-of-the-art EIF methods by a sizable margin in both success rate and execution efficiency.

\end{abstract}

% instruction completer
% multi-modal object localizer
\section{Introduction}
\label{sec:intro}
% ~\cite
%-------------------------------------------------------------------------

\begin{figure}[t]
    \centering
    \includegraphics[width=\columnwidth]{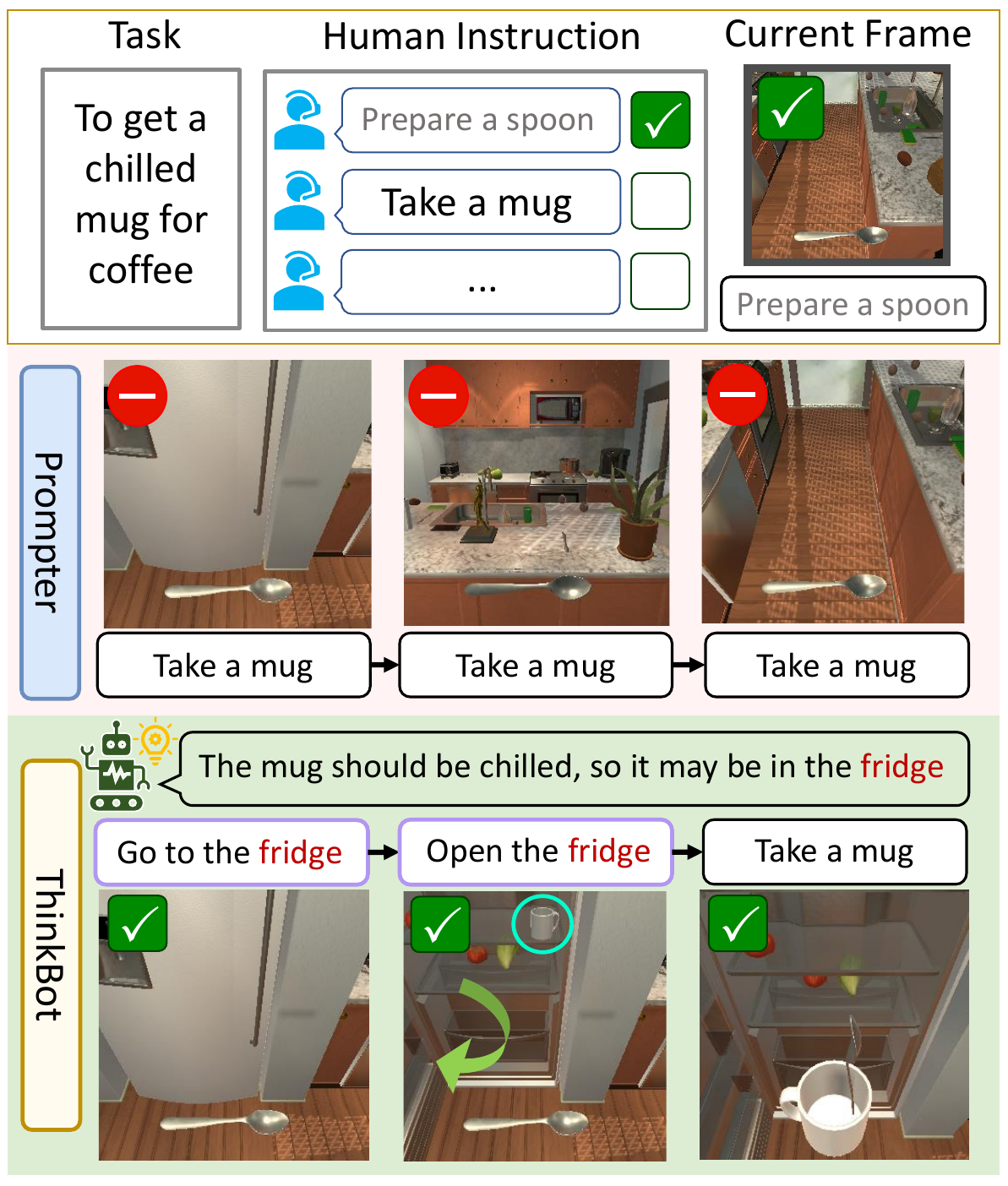}
    \caption{Comparison between conventional EIF methods (Prompter \cite{inoue2022prompter}) and our \method. Existing methods directly leverage sparse human instruction to generate action sequence, which usually get stuck due to the incoherence of instruction. Our \method recovers missing action descriptions by reasoning the thought chain in sparse human instruction, and can successfully complete challenging tasks.
    }
    \label{fig:teaser}
    \vspace{-0.4cm}
\end{figure}

% Take a deep breath, I want you to act as an experienced academician. I'm writing a scientific paper for CVPR conference. Here is one part of my draft, please revise it to make it more fluent and compatible with CVPR conference. Only use simple words and expressions. This is very important to my career. """ """

%%% HINT: EIF task importance (esp. instruction importance)
Designing autonomous agents for diverse household tasks has been highly desired in research of artificial intelligence for a long time. 
Recent advances in computer vision \cite{wang2023internimage,li2023blip} and natural language processing \cite{brown2020language,ouyang2022training} enable autonomous agents to complete complex human requirements, because the appeared large pre-trained models can comprehend human instruction and perceive the world accurately.
% more fluent ?
Embodied instruction following (EIF) \cite{misra2017mapping,zhu2017visual,gordon2018iqa,shridhar2020alfred} requires the agent to ground human instruction to consecutive task plans with feasible execution, which requires high success rate and completion efficiency.
%which requires high generalization ability and success rate.

\begin{figure*}[t]
    \centering
    \includegraphics[width=\linewidth]{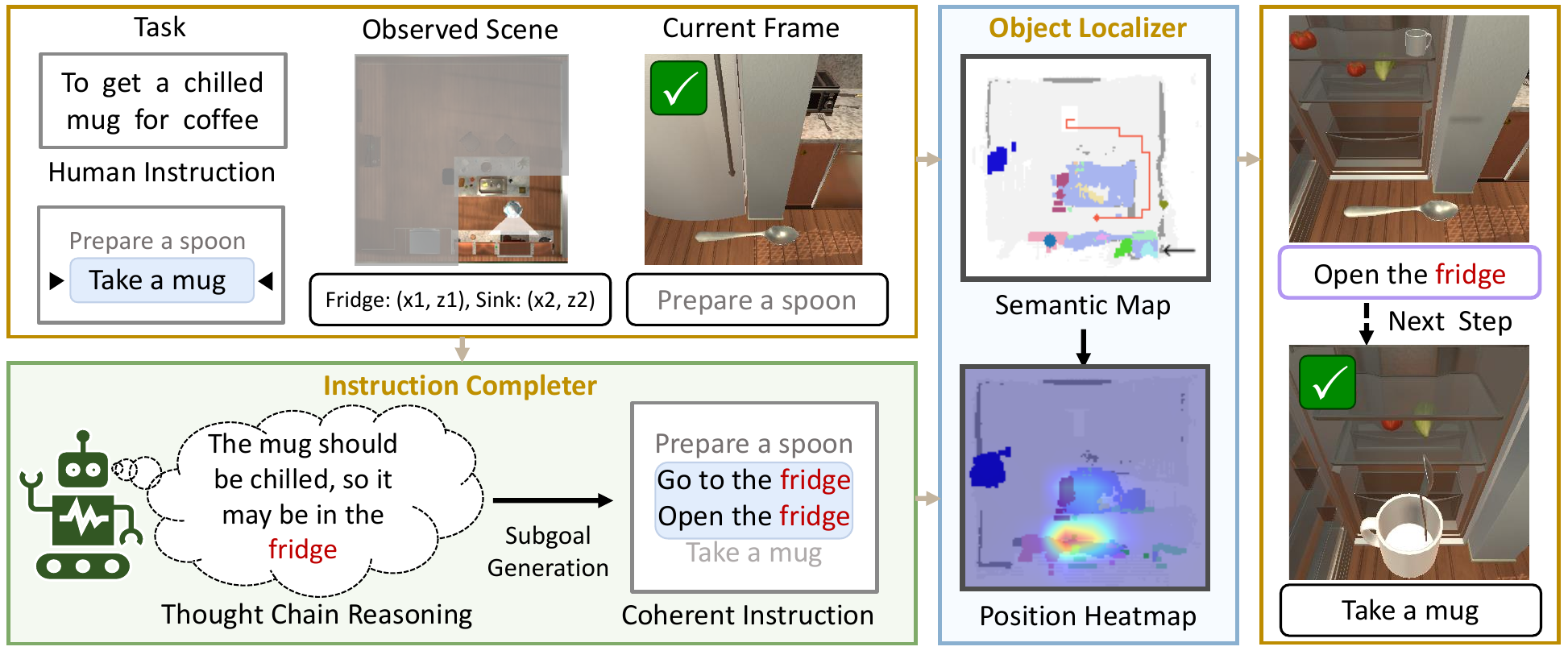}
    % \vspace{-0.5em}
    \caption{ The overall pipeline of \method, which consists of an instruction completer and an object localizer. The instruction completer generates the coherent instruction with interacted objects based on sparse human instruction and the current visual perception results, and the object localizer predicts the position of the interacted object for manipulation and navigation.
        } 
    \label{fig:pipeline}
    \vspace{-0.5em}
\end{figure*}

To accomplish the challenging EIF task, imitation learning \cite{shridhar2020alfred, pashevich2021episodic, singh2021factorizing, song2022one, nguyen2021look, zhang2021hierarchical, nguyen2020hierarchical, suglia2021embodied, kim2021agent} is widely adopted to generate low-level actions from historical observation and given instruction, while they fail to adapt to new scenarios due to the insufficient pair-wise data between human instruction and low-level actions. 
To this end, modular methods \cite{blukis2021persistent, min2022film, murray2022following, inoue2022prompter, kim2023context, ding2023embodied, liu2022planning} decompose complex tasks to high-level planning conditioned on instruction and low-level execution with predefined controllers.
However, human instruction is usually sparse with incoherence for action plan generation of agents. 
%For example, humans may give the instruction \emph{`Slice the bread, put the lettuce on the bread'} for sandwich making. In the realistic scenarios, the lettuce may be stored in the fridge and should be sliced before put on the bread. Therefore, the dense instruction should be \emph{`Slice the bread, take the lettuce from the fridge, slice the lettuce, put the lettuce on the bread'}. 
% For example, in \cref{fig:teaser}, humans may give the instruction \emph{`Prepare a spoon, Take a mug'} for coffee making. 
For example, in \cref{fig:teaser}, humans may give the instruction \emph{`Prepare a spoon, Take a mug'} for making cold brew coffee.
In realistic scenarios, the mug may be stored in a fridge, and the agent may need to open the fridge first to get the mug. Therefore, the dense instruction should be \emph{`Go to the fridge, open the fridge, and take a mug'}. 
Lacking coherent instruction usually disables the agent to acquire feasible action sequences in execution, and the success rate across diverse tasks in complicated indoor environments still remains low.

In this paper, we propose a \method agent to accurately complete diverse EIF tasks in interactive environments. 
Unlike existing methods that directly employ the sparse human instruction for agent action sequence generation, we recover the missing action descriptions for agent execution by reasoning the thought chain in sparse human instruction. Therefore, coherent instruction can be leveraged to generate more feasible agent actions in complex EIF tasks especially with long sequences. 
More specifically, we first propose an instruction completer based on large language models to predict the missing actions with interacted objects in the sparse human instruction, where we carefully design prompts to consider the perceived objects in the surrounding environments and the completed sub-goals in the task execution sequences. 
We then present a multimodal object localizer that predicts the position of the recovered missing objects based on the scene semantic maps, where the mined object correlation is also leveraged to enhance the localization ability.
Extensive experiments on ALFRED \cite{shridhar2020alfred} show that our \method outperforms the state-of-the-art EIF methods by a sizable margin in both success rate and execution efficiency. Our main contributions can be summarized as follows:

\begin{itemize}
    \item We propose a \method agent that reasons the thought chain in sparse human instruction for coherence mining to successfully complete complex EIF goals.

    \item  We present an instruction completer based on large language models to generate the missing actions with interacted objects, and propose an object localizer to predict the position of objects for interaction.

    \item We conduct extensive experiments of diverse EIF tasks on ALFRED benchmark, and the results demonstrate that our method achieves higher success rate and path-length-weighted success rate than the state-of-the-art methods on unseen environments.
    
\end{itemize}

\section{Related Work}
\label{sec:related_work}

% HINT: EIF importance
\textbf{Embodied Instruction Following:} Developing generalist agents that can follow human instruction to complete diverse tasks in interactive environments is a long-standing goal.
In the pursuit of this goal, EIF has been widely studied in recent years for high generalizability and practicality.
Prior arts can be divided into two categories: end-to-end methods and modular methods.
% HINT: end-to-end methods
End-to-end methods directly generate low-level actions conditioned on the current state of the environment and human instruction \cite{shridhar2020alfred, pashevich2021episodic, singh2021factorizing, song2022one, nguyen2021look, zhang2021hierarchical, nguyen2020hierarchical, suglia2021embodied, kim2021agent}.
For instance, Pashevich \etal \cite{pashevich2021episodic} developed an episodic transformer to encode language inputs and the episode history of visual observation and actions, which was decoded for action sequence generation with auto-regression. 
However, end-to-end methods often struggle to generalize to unseen scenes due to insufficient pair-wise data between instruction and low-level action sequences. 
To address this, modular methods \cite{blukis2021persistent, min2022film, murray2022following, inoue2022prompter, kim2023context, ding2023embodied, liu2022planning} plan high-level action sequences and execute them with pre-defined local policies guided by online semantic maps, which are free of the pair-wise data between instruction and low-level actions. 
In modular methods, selecting correct targets and actions for navigation and interaction is very important for searching efficiency and task success. Min \etal \cite{min2022film} directly employed convolutional networks to predict the target position from current semantic map, and Murray \etal \cite{murray2022following} proposed a landmark classification model based on BERT \cite{kenton2019bert} for target selection only based on original human instruction. Inoue \etal \cite{inoue2022prompter} predicted landmark objects according to offline co-occurrence probabilities of objects evaluated by pre-trained language models, and Kim \etal \cite{kim2023context} yield the detailed plan by incorporating the contextual information of natural language instructions.
%and Kim \etal \cite{kim2023context} leveraged the object correlation information to yield the detailed plan with semantic-rich context. 
Nevertheless, directly considering the sparse human instruction for agent action generation usually fails to achieve human goals due to the instruction incoherence with missing action descriptions.

% HINT: LLM is important for EAI
\noindent\textbf{LLMs for Embodied Agents:} Large language models (LLMs) \cite{brown2020language,ouyang2022training} have demonstrated their capability in embodied AI tasks such as visual-language navigation \cite{qiao2023march, long2023discuss, shah2023lm,zhou2023navgpt,georgakis2022cross,chen2022weakly}, object navigation \cite{yu2023l3mvn, zhou2023esc}, open-world exploration \cite{wang2023describe, zhu2023ghost}, and embodied planning \cite{mu2023embodiedgpt, yao2022react, wu2023embodied,brohan2023can,huang2022language,huanginner,raman2022planning,singh2023progprompt,lu2022neuro}, where the high generalizability across deployment scenes and downstream tasks of LLMs enables embodied agents to achieve diverse human goals in complex environments. 
With the rich commonsense embedded in LLMs, fine-grained actions regarding human instruction can be directly generated. 
Zhou \etal \cite{zhou2023navgpt} directly prompted LLMs to perform zero-shot sequential action prediction by taking the textual descriptions of historical visual observations as inputs.
To facilitate efficient exploration, ESC~\cite{zhou2023esc} generated frontier candidates on the observed semantic map, and employed LLM to determine the next frontier by considering hand-crafted soft constraints jointly. 
% In addition to determining the target, Qiao \etal \cite{qiao2023march} use LLM to generate fine-grained instruction for a marching agent.
While direct generation of fine-grained actions is challenging for LLMs due to the extremely large search space, other methods decompose the overall solution into high-level plan generation and low-level action controlling \cite{mu2023embodiedgpt, yao2022react, wu2023embodied,brohan2023can,huang2022language,huanginner}. 
Wu \etal \cite{wu2023embodied} crafted a large-scale embodied planning dataset and finetuned different plan generators on the dataset for task plan grounding. Yao \etal \cite{yao2022react} generated consecutive plans by prompting LLMs to synergize reasoning and acting, and achieved impressive performance on text-based benchmarks \cite{ALFWorld20}.
However, despite the high reasoning ability of LLMs, the spatial localization ability for interacted objects is weak in LLMs. We present a multimodal transformer-based object localizer to provide spatial guidance for agents in object interaction accurately.

\section{Approach}
\label{sec:approach}

In this section, we first briefly introduce the problem in EIF and our overall pipeline of the \method agent (\cref{subsec:promblem_statement}). Then we detail the instruction completer that recovers the missing actions with interacted objects for coherence mining (\cref{subsec:instruction_completer}), and demonstrate the multimodal object localizer which provides spatial guidance for agents in interaction (\cref{subsec:object_localizer}).

%An embodied instruction following agent performs a sequence of navigational steps and object interactions based on egocentric visual observations it receives from the environment. These actions and interactions are based on natural language instruction that the agent must follow to accomplish the task. 

\subsection{Problem Statement and Overall Pipeline}
\label{subsec:promblem_statement}
The embodied instruction following task requires an agent in the interactive environment to finish the human goals physically by generating action sequences, where the human goals and the step-by-step instruction are given to the agent for guidance. Following the modular methods, we first generate high-level subgoal sequences and then execute them with a pre-defined controller guided by online semantic maps. In the $t_{th}$ step, the agent needs to generate a high-level subgoal $A_t$ based on the current instruction $I_t$ and the object state $S_{t-1}$ after implementing the last subgoal by the controller. A high-level plan $A_t$ is a tuple $(a_t, o_t, p_t)$, where $a_t$ is the primitive action and $o_t $ means the interacted object with the position $p_t$. In EIF tasks, human instruction is usually sparse with significant incoherence between consecutive steps. Therefore, it is very challenging to generate feasible action $A_t$ based on the current step-wise instruction $I_t$ and the object state $S_{t-1}$ after implementing the last action. 
% For example, humans may provide the instruction \emph{`Slice the bread, put the lettuce on the bread'}. Lettuce may be stored in the fridge and can only be put on the bread after being sliced in realistic scenes. Consequently, it is difficult for the agent to put the lettuce on the bread without other instruction after slicing the bread. The coherent instruction should be \emph{`Slice the bread, take the lettuce from the fridge, slice the lettuce, put the lettuce on the bread'}. 
% For example, humans may provide the instruction \emph{`Prepare a spoon, take a mug'} for making cold brew coffee. 
For example, humans may provide the instruction \emph{`Prepare a spoon, take a mug'} for coffee making. 
However, the mug may be stored in the fridge in realistic scenes. Consequently, it is difficult for the agent to take the mug with spoon without other instruction after preparing the bread. The coherent instruction should be \emph{`Prepare a spoon, go to the fridge, open the fridge, take a mug'}. 
Therefore, our goal is to recover the missing action descriptions in the sparse human instruction.

% HINT: reasoning for instruction recovery
Since the thought chain reveals a series of intermediate steps from the initial problem to the final solution \cite{wei2022chain}, it can be leveraged to decompose the original complex problems and enhance the feasibility of the solution. In EIF tasks, reasoning the thought chain can predict the missing action descriptions in the sparse human instruction to successfully achieve the goal. The overall pipeline of our \method agent is shown in \Cref{fig:pipeline}, which consists of an instruction completer recovering the missing actions with interacted objects and an object localizer predicting the object location for agent interaction. For the instruction completer, we design prompts for the pre-trained large language model including the descriptions of scene information and the task completion process, which is expected to reason the thought chain in sparse human instruction to provide coherent instruction. For the object localizer, the generated missing actions with interacted objects and the perceived semantic map of the scene are utilized to predict the object location for the agent to interact with, where multimodal transformers are employed for the alignment between language instruction and visual clues in the environment. Finally, the agent can easily complete the human goals to achieve significantly higher success rate with the coherent instruction and the explicit interaction location. 

%Moreover, we learn the local graph attention to impose the object connectivity priors to improve the object localization in the large indoor environment. 

%Landmark reasoning requires global scene-level understanding of the visual observation to abstract it to a resulting action.

\subsection{Instruction Completer}
\label{subsec:instruction_completer}
\begin{figure}[t]
    \centering
    \includegraphics[width=\linewidth]{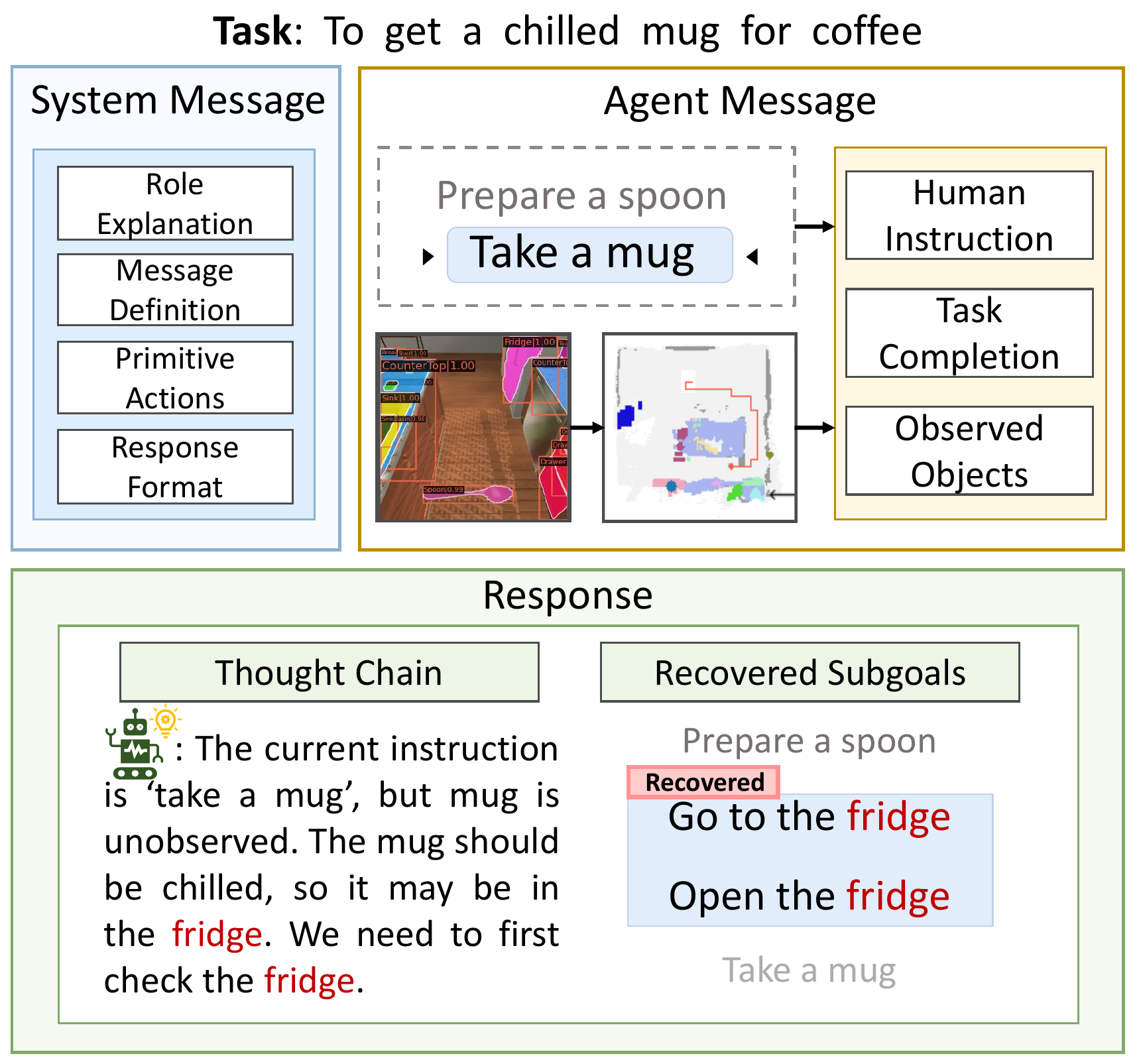}
    % \vspace{-0.5em}
    \caption{ Input and output of the instruction completer based on LLMs. The input contains system message describing the world properties and agent message demonstrating perceived environment information. The output includes the thought chain in sparse human instruction and missing subgoals with interacted objects.
        } 
    \label{fig:instruction_completer}
    \vspace{-0.3cm}
\end{figure}

% HINT: Motivation
To recover the missing action descriptions in the sparse human instruction, we employ LLMs with rich commonsense to reason the thought chain in the instruction.
%The generation process of our instruction completer can be formulated as follows:
%\begin{equation}
    %(a_t, o_t)=\Phi(I_t, S_{t}, G, A_{t-1}),
%\end{equation}
%where $S_t$ means the observed object state of the scene in the $t_{th}$ step, and $\Phi$ represents our LLM-based instruction completer to generate subsequent action sequences.
% Preliminary: prompt template and few-shot examples
While LLMs have demonstrated remarkable abilities in various tasks, unlocking their complete potential requires prompt engineering. To enable the pre-trained LLMs to predict the missing actions with interacted objects accurately, we carefully design the prompt by organizing the system message and agent message that respectively describe the innate world properties and the perceived information. The input and output of the instruction completer are shown in \Cref{fig:instruction_completer} with an example.

\noindent\textbf{System Message:} The system message describes the innate world properties in the simulator including the role explanation, message definition, primitive actions and response format, which remains unchanged during the whole EIF process for the given task and environment. The role explanation specifies the simulation scenario of household AI assistant for LLM response, and the message definition explains the meaning of each input message. Primitive actions constrain the available selection for agent in manipulation and navigation, while response format limits the response structure for LLMs to generate the missing actions with interacted objects.

\noindent\textbf{Agent Message:} The agent message demonstrates the perceived information including the human instruction, task completion process, and observed objects in the scene, which is updated along with the EIF process. The human instruction means the given sparse step-by-step instruction sentences and the final goal. The task completion depicts completed subgoals, the current subgoal and all subgoals, which reflects the execution process of the final goal. The information on observed objects in the scene is represented by the object category and the coordination of the instances that have been seen by the agent, which demonstrates the affordable candidates for interaction during the EIF process. Since providing a large number of candidate objects causes severe hallucination, we only leverage objects that are possible to appear in the current room type as candidate objects. Finally, the system message and the agent message are concatenated to form the prompt for recovering missing actions with interacted objects in the sparse human instruction. 

%A subtask monitor tracks the completion of subtasks for success detection, which provides neccessary guidance to localization in the overall task completion process.

%The observed objects in the scene are the recorded landmarks on the online built semantic map, which is used to provide the affordance candidates for interaction or navigation.
%Besides, while primitive actions can be simply listed in the system template, due to the limited context of LLM, providing all candidate objects may cause severe hallucination, a common problem in LLMs. To address this, we collect and provide possible landmarks in current room type (\eg \emph{`Kitchen'}) as candidate objects for the agent to interact with.

%The feedback contains the last response of the agent and environmental feedback, which is used to self-correct the generated subgoals.
%The system message and the agent message are concatenated to form the prompt. 

\noindent\textbf{Thought Chain Reasoning:} To accurately recover the missing actions in sparse human instruction, we enforce LLMs to reason the thought chain in human instruction that indicates the detailed process from the initial state to the final goal. However, the recovered actions can only be executed if the agent knows the exact position of the interacted objects. Therefore, we also require LLMs to generate the missing subgoals in a structured format for subsequent location prediction. The recovered subgoals are leveraged in subsequent object localizer for position prediction, so that the agent can manipulate the object to successfully achieve the subgoal for task completion.

%For example, the thought \emph{`cut the lettuce in the fridge'} should be parsed into the following subgoals \emph{`Go to the fridge, open the fridge, and slice the lettuce in the fridge'} with the object list \emph{`fridge, lettuce'}. Finally, our instruction completer also generate the expectation for each generated subgoal to detect the action success for task completion process evaluation.

%%%%%%%%%%%%%%%%%%%%%%%%%%%%%%%%%%%%%%%%

\subsection{Multimodal Object Localizer}
\label{subsec:object_localizer}

\begin{figure}[t]
    \centering
    \includegraphics[width=\linewidth]{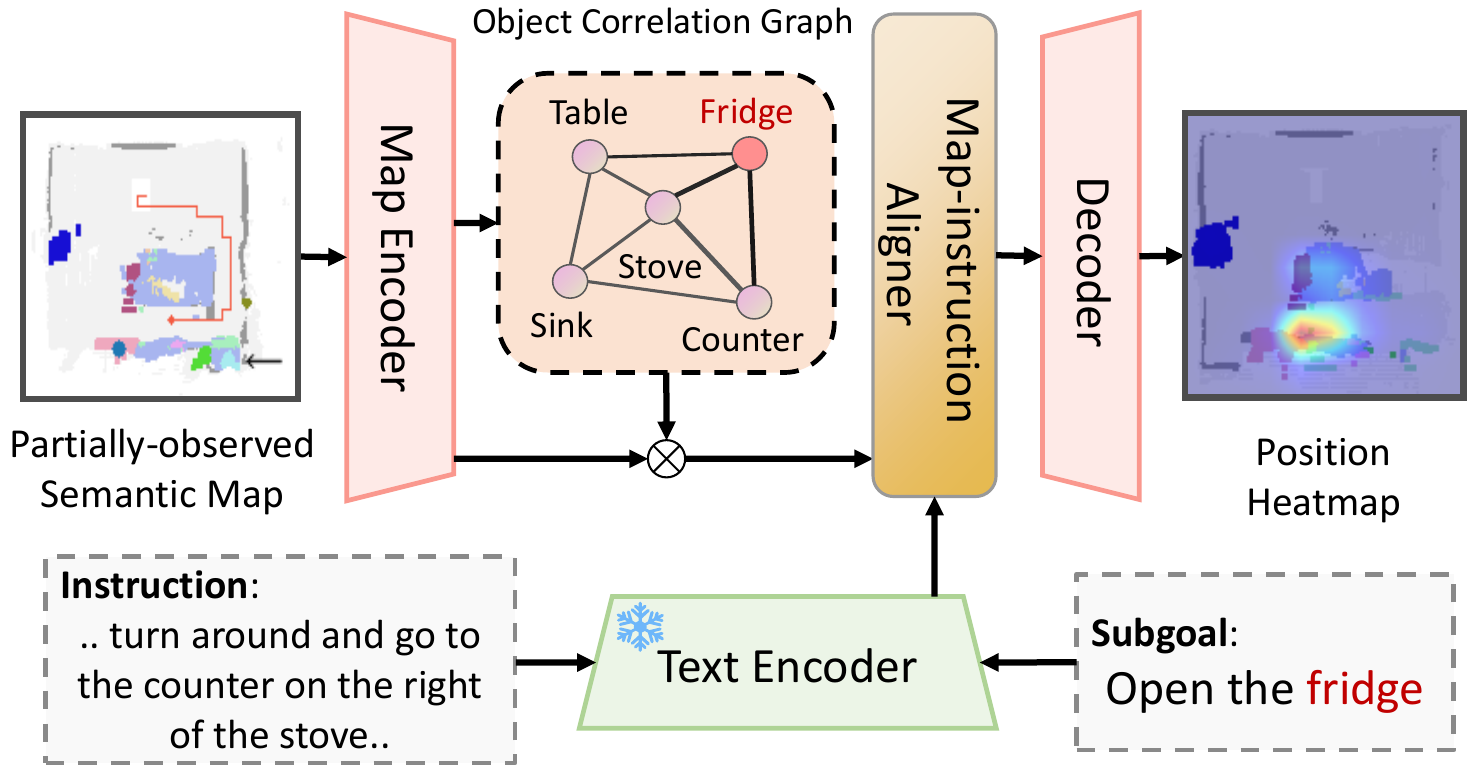}
    % \vspace{-0.5em}
    \caption{The overall pipeline of the multimodal object localizer, which uses recovered instruction and observed semantic map to predict object positions for interaction. The object correlation graph is also learned to strengthen the map features.
        }
    \label{fig:multi_modal_object_localizer}
    \vspace{-0.3cm}
\end{figure}

% HINT: Motivation
Although LLMs can accurately reason the thought chain in sparse human instruction to recover the missing actions with interacted objects, the inferring of object position for agent interaction remains challenging because of the weak spatial localization ability of the language description. To deal with the challenge, we present an object localizer based on multimodal transformers to predict the position of interacted objects based on the recovered instruction and the partially observed semantic map. \Cref{fig:multi_modal_object_localizer} depicts the framework of the object localizer. The recovered instruction acquired from the instruction completer provides the goal for the agent to interact with, and the partially observed semantic map demonstrates the positions of seen objects. The representation of semantic maps and the instruction are leveraged by the decoder to generate the probability heatmap for location prediction of the interacted object. Since the semantic correlation among objects can provide beneficial priors to precisely locate the interacted objects based on the visual perception results, we mine the object correlation graph to strengthen the semantic map representation. In this section, we first detail the encoder design for all inputs in representation extraction, and then illustrate the alignment strategy among different inputs with the multimodal transformer. Finally, we introduce the learning objective and the training pipelines of the object localizer.

For the instruction encoder, we leverage a pre-trained BERT~\cite{kenton2019bert} to extract the instruction features $\mathbf{X}_s$ for the human instruction and the goal object, which are acquired from the prediction results of instruction completer. All layers except the last one are frozen during the object localizer finetuning for high generalizability in understanding instruction for complex tasks. For the map encoder, we employ convolutional neural networks to extract the initial map features $\mathbf{X}'_t$ that take the partially observed semantic maps as the input. 
Since semantic correlation among objects provides beneficial priors for accurate localization, we mine the object correlation graph to embed the priors into semantic map features for further enhancement of object position prediction. The graph is defined as $\mathcal{G}=(\mathbf{V}, \mathbf{E})$, where $\mathbf{V}$ is the set of nodes and $\mathbf{E}$ are the edges between different nodes. Each node represents a possible object category in the simulator, and the element in the $i_{th}$ row and $j_{th}$ column of $\mathbf{E}$ indicates the correlation of objects between the $i_{th}$ and the $j_{th}$ classes.
% The element $e_{ij}$ is acquired by 
% We then rescale the pair-wise distance between objects with Gaussian kernels as $e_{ij} = \exp (- d_{ij}^2/(2\sigma^2))$. We utilize the length of the shortest path between objects for the pair-wise distance $d_{ij}$, which reveals the spatial connectivity priors. Meanwhile, rescaling the distance with Gaussian kernels can impose more importance on neighbored objects in representation extraction, and the close objects provide sufficient semantic priors than faraway ones.
While conventional graph convolutional networks use external priors to handcraft a predefined graph structure, we construct the object correlation graph by learning from the semantic map features $\mathbf{X}'_t$ to adapt to different scenes.
The object correlation graph is generated from the semantic map features $\mathbf{X}'_t$ by $\mathbf{E}_t = f(\mathbf{X}'_t \mathbf{W}_e)$, where $f(\cdot)$ denotes the activation function and $\mathbf{W}_e$ is a learnable matrix for graph generation.
The process of learning the object correlation graph can be regarded as encoding the object correlation priors among the categories.
We then employ graph convolutional layers to enhance the semantic map representation with the object correlation encoding, and acquire the map features for multimodal alignment according to the following:
% \begin{equation}
%     \mathbf{X}_t = \mathbf{X}'_t + \alpha f( \mathbf{E} \mathbf{X}'_t \mathbf{W}_a),
% \end{equation}
\begin{equation}
    \mathbf{X}_t = \mathbf{X}'_t + \mathbf{E} \mathbf{X}'_t \mathbf{W}_a,
\end{equation}
where $\mathbf{W}_a$ is a learnable weight matrix for message passing to embed the object correlation priors into the map features. %and $\alpha$ is a hyperparameter to indicate the importance of the object correlation priors.

After the representation extraction for the instruction and the semantic map, we leverage a map-instruction aligner to align the instruction features and the map features for object position prediction.
Since the semantic map is updated in an online manner with high frequency during agent navigation, we take the features of the semantic maps as the query. The instruction remains unchanged in a subtask, which is utilized as the key and the value features. Specifically, they can be acquired via the following equations: 
\begin{equation} \label{eq:matrices}
    \mathbf{Q}_s = \mathbf{X}_s \mathbf{W}_q, \quad \mathbf{K}_t = \mathbf{X}_t \mathbf{W}_k, \quad \mathbf{V}_t = \mathbf{X}_t \mathbf{W}_v,
\end{equation}where $\mathbf{Q}_s$, $\mathbf{K}_t$ and $\mathbf{V}_t$ respectively mean the query features from the semantic maps, the key and value features from the instruction. The matrices $\mathbf{W}_q$, $\mathbf{W}_k$, $\mathbf{W}_v$ are leveraged to obtain the query, key and value features respectively. The representation $\mathbf{H}_t^s$ that aligns the visual information provided by semantic maps and the language information in the instruction can be acquired as follows:
\begin{equation}
    \mathbf{H}_t^s = {\rm Softmax} \left( \frac{\mathbf{Q}_s \mathbf{K}_t^T}{\sqrt{d}}\right) \mathbf{V}_t,
\end{equation}
where $d$ means the dimension of the representation $\mathbf{H}_t^s$.
%, and $\alpha$ is a hyperparameter to indicate the importance of the connectivity priors. 
The representation is then leveraged to decode the predicted location of the interacted objects. %with transpose convolutional layers.

% TODO: revise loss description
% In order to train the encoders and the decoder in the object localizer, we need to acquire the groundtruth instance position for the interacted objects that are predicted by the instruction completer. We replay the expert demonstration in the training data, and record the position of interacted objects when the subtask is successfully completed. With the projection to the top-down view, the groundtruth positions are used to supervise the object localizer to predict the position of interacted objects. The model is trained with a pixel-wise binary cross-entropy loss, where the labels of pixels covered by the interacted objects are set to one and otherwise to zero. 
%Finally, xxxxxxxx.

In order to train the encoder, the map-instruction aligner, and the decoder in the object localizer, we need to acquire the groundtruth instance position for the interacted objects that are predicted by the instruction completer. We replay expert demonstrations in the training data, and record the masks of interacted objects when a subtask is successfully completed. With the projection from egocentric mask to the top-down view by depth estimation, the groundtruth positions are used to supervise the object localizer to predict the position of interacted objects. The model is trained with a pixel-wise binary cross-entropy loss, where the labels of pixels covered by the interacted objects are set to one and otherwise to zero.

\begin{table*}
    \centering
    % \small
    \footnotesize
     \caption{Comparison with the state-of-the-art methods in SR, GC, PLWSR, PLWGC on the test seen and test unseen splits.} %and the human performance is also given for reference.}
     \vspace{-0.2cm}
    \renewcommand{\arraystretch}{0.95}
    \setlength\tabcolsep{14pt}
    \begin{tabular}{l *{4}{>{\columncolor{Gray!25}}c} *{4}{c}}
    \toprule
    \multirow{2}{*}{\textbf{Method}} & \multicolumn{4}{c}{\textbf{Test Seen}} & \multicolumn{4}{c}{\textbf{Test Unseen}} \\
    \cmidrule(lr){2-5} \cmidrule(lr){6-9}
    & PLWGC & GC & PLWSR & SR & PLWGC & GC & PLWSR & SR \\
    \midrule
    % Seq2seq (Shridhar \etal \cite{shridhar2020alfred}) & 6.27 & 9.42 & 2.02 & 3.98 & 4.26 & 7.03 & 0.08 & 3.9 \\
    % MOCA (Singh \etal \cite{singh2021factorizing}) & 22.05 & 28.29 & 15.10 & 22.05 & 9.99 & 14.28 & 2.72 & 5.30 \\
    % E.T. (Pashevich \etal \cite{pashevich2021episodic}) & 34.93 & 45.44 & 27.78 & 38.42 & 11.46 & 18.56 & 4.10 & 8.57 \\
    % LWIT (Nguyen \etal \cite{nguyen2021look}) & 23.10 & 40.53 & 43.10 & 30.92 & 16.34 & 20.91 & 5.60 & 9.42 \\
    % HITUT (Zhang \etal \cite{zhang2021hierarchical}) & 17.41 & 29.97 & 11.10 & 21.27 & 11.51 & 20.31 & 5.86 & 13.87 \\
    % ABP (Kim \etal \cite{kim2021agent}) & 4.92 & 51.13 & 3.88 & 44.55 & 2.22 & 24.76 & 1.08 & 15.43 \\
    % \midrule
    % LLM-Planner (Song \etal \cite{song2023llm}) & - & 26.77 & - & 18.20 & - & 23.37 & - & 16.42 \\
    % FILM (Min \etal \cite{min2022film}) & 15.59 & 39.55 & 11.27 & 28.83 & 15.13 & 38.52 & 11.32 & 27.80 \\
    % LGS-RPA (Murray \etal \cite{murray2022following}) & 28.97 & 48.66 & 21.28 & 40.05 & 22.76 & 45.24 & 22.76 & 35.41 \\
    % Prompter (Inoue \etal \cite{inoue2022prompter}) & 30.72 & 63.43 & 25.81 & 53.23 & 26.22 & 58.76 & 20.76 & 45.72 \\
    % CPEM (Kim \etal \cite{kim2023context}) & 27.49 & 59.40 & 22.61 & 50.62 & 27.00 & 61.10 & 22.61 & 49.84 \\
    Seq2seq & 6.27 & 9.42 & 2.02 & 3.98 & 4.26 & 7.03 & 0.08 & 3.9 \\
    MOCA & 22.05 & 28.29 & 15.10 & 22.05 & 9.99 & 14.28 & 2.72 & 5.30 \\
    E.T. & 34.93 & 45.44 & 27.78 & 38.42 & 11.46 & 18.56 & 4.10 & 8.57 \\
    LWIT & 23.10 & 40.53 & 43.10 & 30.92 & 16.34 & 20.91 & 5.60 & 9.42 \\
    HITUT & 17.41 & 29.97 & 11.10 & 21.27 & 11.51 & 20.31 & 5.86 & 13.87 \\
    ABP & 4.92 & 51.13 & 3.88 & 44.55 & 2.22 & 24.76 & 1.08 & 15.43 \\
    \midrule
    LLM-Planner & - & 26.77 & - & 18.20 & - & 23.37 & - & 16.42 \\
    FILM & 15.59 & 39.55 & 11.27 & 28.83 & 15.13 & 38.52 & 11.32 & 27.80 \\
    LGS-RPA & 28.97 & 48.66 & 21.28 & 40.05 & 22.76 & 45.24 & 22.76 & 35.41 \\
    Prompter & 30.72 & 63.43 & 25.81 & 53.23 & 26.22 & 58.76 & 20.76 & 45.72 \\
    CPEM & 27.49 & 59.40 & 22.61 & 50.62 & 27.00 & 61.10 & 22.61 & 49.84 \\
    % Prompter+ (Ours) & 37.07 & 69.55 & 32.96 & 61.90 & 30.44 & 67.43 & 26.87 & 57.36 \\
    % Prompter+ (Ours) & 37.07 & 69.55 & 32.96 & 61.90 & 30.59 & 66.97 & 26.75 & 57.03 \\
    % Prompter+ (Ours) & 36.35 & 70.20 & 31.12 & 60.86 & 30.59 & 66.97 & 26.75 & 57.03 \\
    Prompter+ & 36.35 & 70.20 & 31.12 & 60.86 & 30.09 & 65.71 & 26.22 & 55.46 \\
    \midrule
    % \textbf{\method} (Ours) &  &  &  &  & 30.71 & 67.54 & 26.95 & 57.68 \\
    \textbf{\method} (Ours) & \textbf{37.01} & \textbf{71.64} & \textbf{32.02} & \textbf{62.69} & \textbf{30.73} & \textbf{67.75} & \textbf{26.93} & \textbf{57.82} \\
    % \midrules
    % Human & - & - & - & - & 87.60 & 94.50 & 85.80 & 91.00 \\
    \bottomrule
    \end{tabular}
    \label{table:comparison_with_sota}
\end{table*}

% compute by normalization
% \begin{table}
% \small
%     \centering
%     \begin{tabular}{lccc}
%     \toprule
%     \textbf{Error mode} & FILM & Prompter & \textbf{\method} \\
%     \midrule
%     Goal object not found & 31.09 & 32.73 &  \\
%     Interaction failures & 10.19 & 18.78 &  \\
%     Collisions & 13.12 & 0.31 &  \\
%     Language processing & 29.27 & 32.8 &  \\
%     Others & 16.33 & 9.6 &  \\
%     \bottomrule
%     \end{tabular}
%     \caption{\textbf{Comparisons of error modes.} Compared with the state-of-the-art method (FILM and Prompter), our \method perform better in dealing with `Goal Object not found' error, which is the dominated error of the counterparts.}
%     \label{table:error_modes}
% \end{table}

\section{Experiments}
\label{sec:experiments}
In this section, we first introduce the experiment setup including datasets, baseline methods, evaluation metrics and implementation details. Then we compare our method with the state-of-the-art EIF approaches to show the superiority in success rate and efficiency, and conduct an ablation study to verify the effectiveness of the instruction completer and the object localizer. Finally, we also demonstrate the visualization results of our method to depict our intuition.
Additional results and case studies are provided in the supplementary material.

\subsection{Experimental Setup}

\textbf{Dataset:} For the simulation of EIF tasks, we utilize the well-recognized ALFRED benchmark \cite{shridhar2020alfred} within the AI2-THOR \cite{kolve2017ai2} virtual environment. The ALFRED benchmark includes \num{25},{743} trajectory-instruction pairs, covering \num{7} different task types with varying levels of complexity. The benchmark is divided into five splits including \emph{train}, \emph{test seen}, \emph{test unseen}, \emph{valid seen} and \emph{valid unseen}. The ALFRED benchmark poses significant challenges for EIF agents, as it requires them to ground incoherent natural instruction of different granularity into various household tasks that involve long-horizon reasoning plans.

\noindent\textbf{Baselines:} We compare our agent, \method, with previously published state-of-the-art EIF models. The counterparts include end-to-end methods Seq2seq \cite{shridhar2020alfred}, MOCA \cite{singh2021factorizing}, E.T. \cite{pashevich2021episodic}, LWIT \cite{nguyen2021look}, HITUT \cite{zhang2021hierarchical}, ABP \cite{kim2021agent}, and modular methods LLM-Planner \cite{song2023llm}, FILM \cite{min2022film}, LGS-RPA \cite{murray2022following}, Prompter \cite{inoue2022prompter}, CPEM \cite{kim2023context}. We also construct a strong baseline in our experiments, which is a modified version of Prompter \cite{inoue2022prompter} denoted as Prompter+. The strong baseline combines environment-aware memory \cite{kim2023context} and a re-trained object detector \cite{wang2023internimage} with the vanilla Prompter.

\noindent\textbf{Evaluation Metrics:} We follow the evaluation protocol outlined in the ALFRED benchmark.
The primary metric is the success rate (SR) that measures the percentage of tasks completed, and we also report the goal-condition success rate (GC), which evaluates the percentage of satisfied goal conditions for all subgoals in step-by-step instruction.
To account for efficiency in task completion, both SR and GC are penalized by the length of the execution sequence to compute a path-length-weighted (PLW) score for each metric, which are termed PLWSR and PLWGC respectively.

\noindent\textbf{Implementation Details:} The instruction completer adopts the publicly released GPT-3.5 API \texttt{GPT-3.5-turbo} as the backbone architecture, where we set the generation temperature to \num{0} for stability enhancement. For prompt design, we leverage emotion prompt \cite{li2023emotionprompt} and prompt optimization \cite{yang2023large} in the system message template to further boost the performance of LLMs. For the multimodal object localizer, we employ a truncated ResNet18 \cite{georgakis2022cross} for the map encoder and follow \cite{vaswani2017attention} to mine the inter-modal representation interaction with the multimodal transformer. The object correlation graph is learned via graph convolutional layers \cite{du2020learning}. Moreover, we collect a dataset of partially-observed semantic maps and target location map pairs by replaying expert trajectories in the simulator for object localizer training, and AdamW optimizer \cite{loshchilov2017decoupled} with the initial learning rate \num{5e-4} and step decay is employed for parameter update.

% original table submission
% \begin{table*}
%     \centering
%     \footnotesize
%     \caption{Comparison of methods combining different proposed techniques, where valid unseen and the selected hard valid unseen splits are used for evaluation.}
%     \renewcommand{\arraystretch}{0.95}
%     \setlength\tabcolsep{12pt}
%     \begin{tabular}{l *{4}{>{\columncolor{Gray!25}}c} *{4}{c}}
%     \toprule
%     \multirow{2}{*}{\textbf{Method}} & \multicolumn{4}{c}{\textbf{Valid Unseen}} & \multicolumn{4}{c}{\textbf{Hard Valid Unseen}} \\
%     \cmidrule(lr){2-5} \cmidrule(lr){6-9}
%     & PLWGC & GC & PLWSR & SR & PLWGC & GC & PLWSR & SR \\
%     \midrule
%     Random & 26.18 & 67.64 & 23.80 & 59.68 & 0.32 & 5.41 & 0 & 0 \\
%     Prompter+ & 29.36 & 72.00 & 26.82 & 64.43 & 0.48 & 5.41 & 0 & 0 \\
%     Groundtruth Location & 39.71 & 72.75 & 37.01 & 67.97 & 0.79 & 5.41 & 0 & 0 \\
%     \midrule
%     w/o Instruction Completer & 29.09 & 72.38 & 26.43 & 64.92 & 0.48 & 5.41 & 0 & 0 \\
%     w/o Object Localizer  & 30.24 & 74.37 & 27.87 & 67.02 & 9.29 & 22.41 & 8.11 & 16.22 \\    % i.e., Prompter+ w/ LLM, random w/ LLM
%     w/o Object Correlation Graph & 30.41 & 73.89 & 28.14 & 67.34 & 11.31 & 29.46 & 9.74 & 21.62 \\    % 
%     \textbf{\method} (Ours) & \textbf{31.11} & \textbf{75.30} & \textbf{28.73} & \textbf{67.77} & \textbf{11.95} & \textbf{30.86} & \textbf{10.26} & \textbf{22.97} \\
%     \bottomrule
%     \end{tabular}
%     \label{table:ablation_study}
% \end{table*}

% revised
\begin{table*}
    \centering
    % \small
    \footnotesize
    \caption{Comparison of methods combining different proposed techniques, where valid unseen and the selected hard valid unseen splits are used for evaluation.}
    % \vspace{-0.2cm}
    \renewcommand{\arraystretch}{0.95}
    \setlength\tabcolsep{12pt}
    \begin{tabular}{l *{4}{>{\columncolor{Gray!25}}c} *{4}{c}}
    \toprule
    \multirow{2}{*}{\textbf{Method}} & \multicolumn{4}{c}{\textbf{Valid Unseen}} & \multicolumn{4}{c}{\textbf{Hard Valid Unseen}} \\
    \cmidrule(lr){2-5} \cmidrule(lr){6-9}
    & PLWGC & GC & PLWSR & SR & PLWGC & GC & PLWSR & SR \\
    \midrule
    Random & 26.18 & 67.64 & 23.80 & 59.68 & 0.32 & 5.41 & 0 & 0 \\
    % w/ FILM & 28.74 & 72.46 & 26.58 & 64.31 & 0.29 & 4.76 & 0 & 0 \\  % supplementary?
    % Prompter+ & 29.63 & 72.70 & 27.19 & 65.04 & 0.48 & 5.41 & 0 & 0 \\
    Prompter+ & 29.36 & 72.00 & 26.82 & 64.43 & 0.48 & 5.41 & 0 & 0 \\
    Groundtruth Location & 39.71 & 72.75 & 37.01 & 67.97 & 0.79 & 5.41 & 0 & 0 \\
    \midrule
    % w/o Instruction Completer & 29.76 & 73.08 & 27.66 & 65.53 & 0.48 & 5.41 & 0 & 0 \\    % i.e., Prompter+ w/ Localizer
    w/o Instruction Completer & 29.09 & 72.38 & 26.43 & 64.92 & 0.48 & 5.41 & 0 & 0 \\
    w/o Object Localizer  & 30.24 & 74.37 & 27.87 & 66.99 & 9.29 & 22.41 & 8.11 & 16.22 \\    % i.e., Prompter+ w/ LLM, random w/ LLM
    w/o Object Correlation Graph & 30.41 & 73.89 & 28.14 & 67.36 & 11.31 & 29.46 & 9.74 & 21.62 \\    % 
    \textbf{\method} (Ours) & \textbf{31.11} & \textbf{75.30} & \textbf{28.73} & \textbf{67.72} & \textbf{11.95} & \textbf{30.86} & \textbf{10.26} & \textbf{22.97} \\
    \bottomrule
    \end{tabular}
    \label{table:ablation_study}
    % \vspace{-0.1cm}
\end{table*}

\begin{figure*}[t!]
    \small
    \centering
    \includegraphics[width=\linewidth]{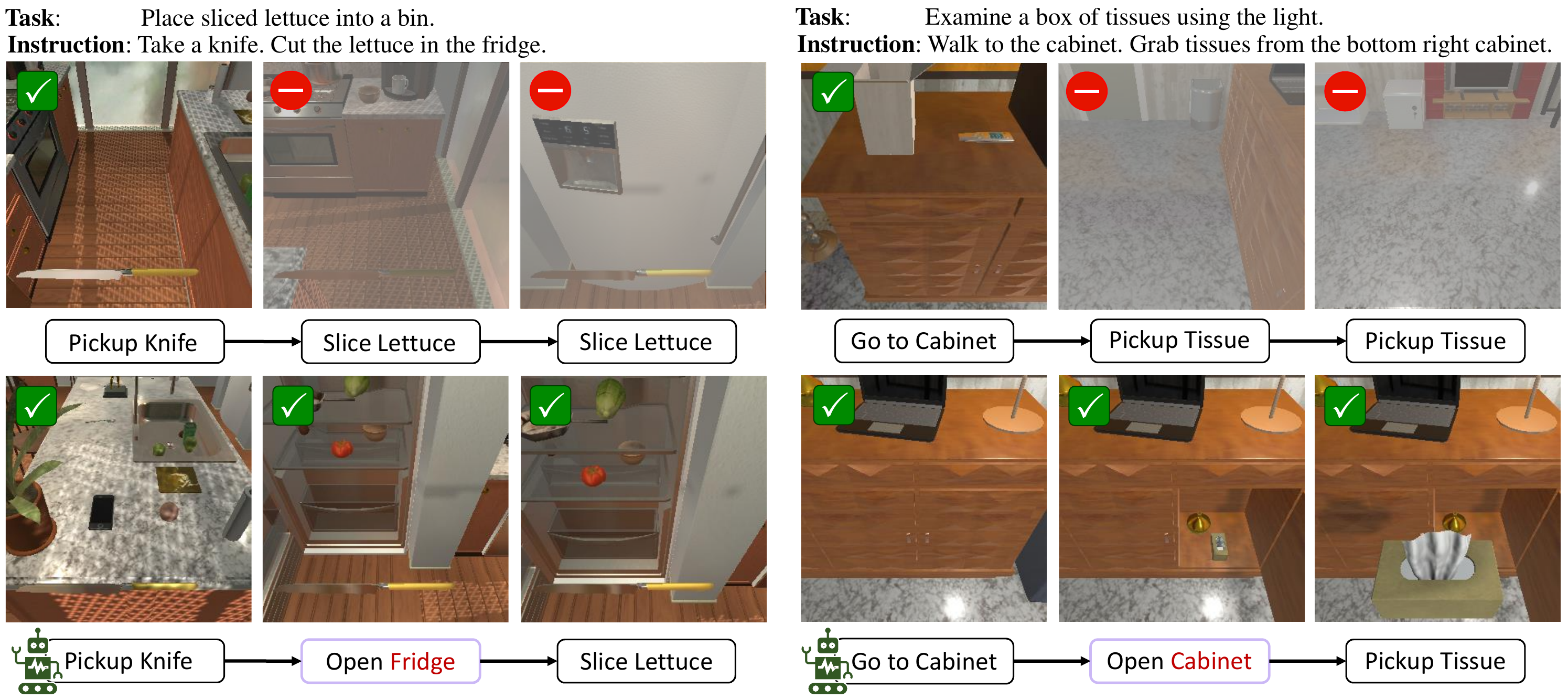}
    % \vspace{-.5em}
    \vspace{-0.2cm}
    \caption{Visualization of the agent action sequence acquired by Prompter+ (top) and our ThinkBot (bottom), where our method can recover the missing actions with interacted instances `Open Fridge' and `Open Cabinet' to successfully achieve the human goal.}
    % \vspace{-1em}
    % \vspace{-0.1cm}
    \label{fig:qualitative_case_study}
\end{figure*}

\subsection{Comparison with the State-of-the-Arts}
In this section, we compare the proposed \method with the state-of-the-art methods on the ALFRED benchmark. \Cref{table:comparison_with_sota} illustrates the comparison of SR, GC, PLWGC, and PLWSR on the test splits for both seen and unseen scenarios. Our \method achieves the best performance on all four metrics in both test seen and test unseen split, outperforming the previous arts including both end-to-end and modular methods by a sizable margin. 
Compared with CPEM \cite{kim2023context}, \method surpasses by 7.98\% and 12.07\% SR on the test unseen and test seen split, respectively.
Besides, Prompter leverages the large language models to infer object co-occurrence probability for semantic search in EIF, and is further enhanced with better visual perception modules to acquire our Prompter+. However, they still suffer from the sparse human instruction with incoherence that usually causes execution failure. 
On the contrary, our \method reasons the thought chain in the sparse human instruction to recover the missing action descriptions, which can provide coherent instruction for the agent to successfully complete the human goal with high efficiency. 
% test seen
As a result, our method outperforms the second-best Prompter+ method by 1.83\% (62.69\% vs. 60.86\%) and 1.44\% (71.64\% vs. 70.20\%) in SR and GC respectively in the test seen split. 
% test unseen
% Moreover, the advantages of our method are more obvious in the test unseen split, which are 0.79\% (57.82\% vs. 57.03\%) in SR and 0.78\% (67.75\% vs. 66.97\%) in GC. 
Moreover, the advantages of our method are more obvious in the test unseen split, which are 2.36\% (57.82\% vs. 55.46\%) in SR and 2.04\% (67.75\% vs. 65.71\%) in GC. 
The results indicate the high generalization ability of our ThinkBot even in novel scenarios that are never seen in training data.
In terms of the efficiency in task completion, our method achieves \num{32.02}\% PLWSR and \num{37.01}\% PLWGC in the test seen split, and \num{26.93}\% PLWSR and \num{30.73}\% PLWGC in the test unseen split, which demonstrates the efficiency of our method. %in task completion. 
% In conclusion, our \method agent is more practical than stat-of-the-art methods for EIF tasks in realistic deployment scenarios with complex environments and long sequences.
In conclusion, our \method agent is more practical than stat-of-the-art methods for EIF tasks in scenarios with complex environments and long sequences.

\begin{figure}[t!]
    \centering
    \includegraphics[width=\columnwidth]{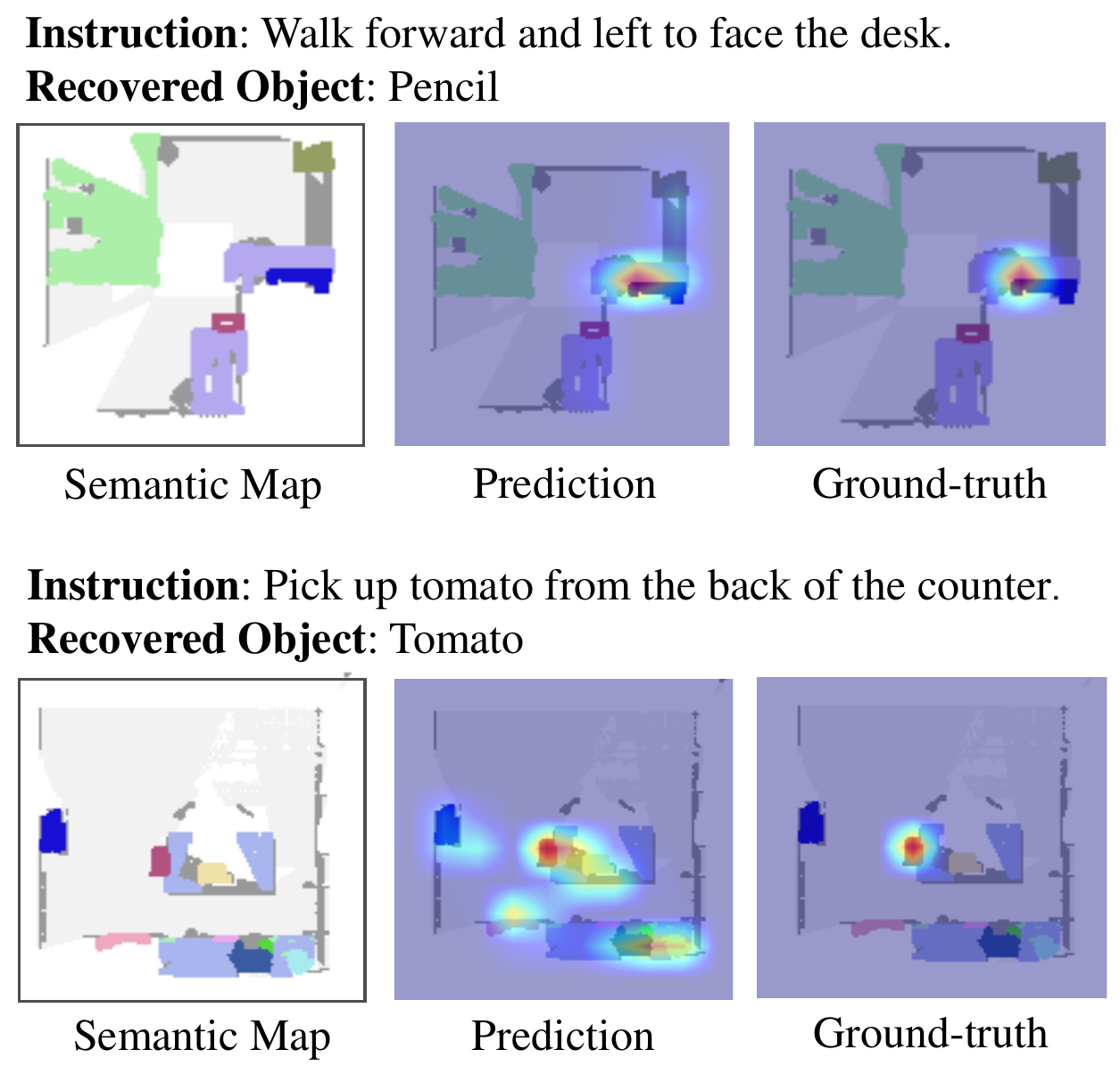}
    \vspace{-0.4cm}
    \caption{ 
        The visualization of the predicted and groundtruth positions of interacted objects, where the partially observed semantic maps are also depicted.}
    \label{fig:qualitative_object_localizer}
    \vspace{-0.3cm}
\end{figure}

\subsection{Ablation Study}
% \paragraph{Ablation study of Different Instruction-following Strategies:}
Our instruction completer recovers the missing actions with interacted objects by reasoning thought chain in sparse human instruction, and the object localizer predicts the position of the interacted objects. 
In \Cref{table:ablation_study}, we conduct an ablation study to verify the effectiveness of each presented technique. Since recovering the incoherent instruction is especially beneficial in the hard cases where target objects are located inside closed containers (e.g. The task \emph{`to get a chilled mug for coffee'} where the mug is in a closed fridge), we also evaluate different methods in the hard cases extracted from valid unseen split, referred to as \emph{hard valid unseen}. In this subset, the agent must open the receptacles to locate the target objects by recovering the missing interaction with the receptacles from sparse instruction, and then the challenging tasks in the hard cases can be successfully completed.

\noindent\textbf{Instruction Completer:} We first implement our \method without the instruction completer, where we directly predict the location of the target in the original sparse human instruction with the object localizer for interaction. 
The results in success rate drops \num{2.24}\% compared with our vanilla \method in the valid unseen split. 
Directly predicting the location of interacted objects in incoherent instruction usually causes large deviation, because the semantic correlation is weak between the target object and the observed semantic map. 
In hard cases, the agent is more likely to fail to complete the task due to the missing interaction with the containers, and the invisible target object is hard to discover by the agent. 
As a result, the performance in SR without instruction completer drops to zero in hard unseen split.
On the contrary, our instruction completer can reason intermediate action descriptions to mine the semantic correlation, which provides fine-grained instruction for the agent to achieve human goals.

\noindent\textbf{Object Localizer:} We also evaluate the performance of our \method without the object localizer, where the agent leverages the baseline Prompter method to predict interaction location with the coherent human instruction recovered by our instruction completer. We observe notable performance drops in SR and PLWSR. 
% w/o object localizer
In the valid unseen split, the performance drops \num{0.75}\% in SR and \num{0.86}\% in PLWSR compared with our vanilla \method.
% w/o object correlation graph
% especially in valid unseen split
Meanwhile, we also implement the object localizer without the object correlation graph, the performance is also degraded in both valid unseen and hard valid unseen split, because the object localizer is required to predict the position of both the target objects and the corresponding containers. The correlation mining between target objects and the containers can significantly enhance localization accuracy.
% random and Prompter+
Besides, we present the results of Prompter+ with random and Prompter's search policy, which drop significantly in SR and PLWSR.
Hence, the agent without the object localizer fails to locate those target objects to achieve the generated intermediate subgoals, which leads to the failure of task completion. 
% compared with groundtruth
We also provide the results for the settings where the groundtruth location of the interacted objects is given. Our \method achieves a similar success rate, which indicates the precise prediction of the object location.
In conclusion, the results verify the necessity of the object localizer for agent interaction in order to achieve human goals. 

\subsection{Qualitative Analysis}

% \quad 
\textbf{Action Sequence Visualization:} 
% \paragraph{Action Sequence Visualization:} 
% We present qualitative examples of the generated action sequence in \Cref{fig:qualitative_case_study} from Prompter+ and our ThinkBot. 
% In this case, the agent is instructed to \emph{`Take a knife. Cut the lettuce in the fridge'}. The results show that an agent without the instruction completer struggles to complete the task due to the missing `open' action and interacted object `fridge' in the sparse human instruction.
% In contrast, our \method first reasons the thought chain of human instruction (\ie \emph{`open the fridge, slice the lettuce'}), and then recovers the missing `open' action and interacted object `fridge' from the instruction, thus successfully completes the task.
% The case study demonstrates the effectiveness of the instruction completer in recovering the missing actions and interacted objects from sparse human instruction.
We present two qualitative examples of the generated action sequence in \Cref{fig:qualitative_case_study} from Prompter+ and our ThinkBot. 
In the left case, the agent is instructed to \emph{`Take a knife. Cut the lettuce in the fridge'}. The results show that the previous agent struggles to complete the task due to the missing `open' action and interacted object `fridge'.
In contrast, our \method first reasons the thought chain of human instruction, and then recovers the missing `open' action and interacted object `fridge' from the instruction, thus successfully completes the task.
In the right case, \method not only recovers the missing `open' action and interacted object `cabinet', but also interacts with the right cabinet that contains the tissue box.
The case study demonstrates the effectiveness of \method in recovering the missing actions and interacted objects from sparse human instruction.
% More visualization results are provided in the supplementary material.

\noindent\textbf{Visualization of the Object Localizer:} 
% online semantic map
\Cref{fig:qualitative_object_localizer} shows the groundtruth and the predicted location of the interacted objects from the object localizer, where the object category is generated from the upstream instruction completer. 
% The distance between the predicted and the groundtruth positions is also demonstrated. 
In the top case, the interacted object `pencil' is located on the table, and the object localizer predicts the position of the interacted object with negligible deviation. In the bottom case where multiple instances exist for the recovered object, the object localizer is able to locate all objects in the same category. Meanwhile, our ThinkBot can correctly assign the largest probability to the instance that is described by the instruction, and the human goals can be successfully achieved following the step-by-step human instruction.

\section{Conclusion}
\label{sec:conclusion}
In this paper, we have presented a ThinkBot agent that reasons the thought chain for missing instruction recovery in EIF tasks. We design an instruction completer to predict the intermediate actions with interacted objects between incoherent human instruction, and then leverage a multimodal transformer to infer the interaction location for the agent. Extensive experiments in a wide variety of EIF tasks demonstrate the superiority of our method regarding the success rate and execution efficiency.
% future work?

% {
%     \small
%     \bibliographystyle{ieeenat_fullname}
%     \bibliography{reference, cpem}
% }

{
    \small
    \bibliographystyle{ieeenat_fullname}
    \bibliography{reference, cpem}
}

\end{document}